\documentclass{article}

\usepackage{arxiv}

\usepackage[utf8]{inputenc} 
\usepackage[T1]{fontenc}    
\usepackage{hyperref}      
\usepackage{url}           
\usepackage{booktabs}      
\usepackage{amsfonts}      
\usepackage{nicefrac}    
\usepackage{microtype}    
\usepackage{lipsum}
\usepackage{graphicx}
\usepackage{wrapfig}
\usepackage[backend=biber, 
    style=apa]{biblatex}
\usepackage{amsmath}
\usepackage[short]{optidef}

\usepackage[toc,page]{appendix}
\usepackage{amsthm}
\usepackage{subcaption}
\DeclareMathOperator{\Tr}{tr}

\usepackage{tikz}
\usetikzlibrary{arrows, automata}
\usetikzlibrary{bayesnet}
\usepackage{dirtytalk}

\title{Slow Feature Analysis as \\Variational Inference Objective}

\author{
 Merlin Schüler\thanks{
  Institute for Neural Computation, Faculty of Computer Science, Ruhr University Bochum,  Germany.\\Corresponding author:\texttt{merlin.schueler@ini.rub.de}}
  \And
 Laurenz Wiskott\footnotemark[1]
}

\begin{document}
\maketitle
\begin{abstract}
This work presents a novel probabilistic interpretation of Slow Feature Analysis (SFA) through the lens of variational inference. 
Unlike prior formulations that recover linear SFA from Gaussian state-space models with linear emissions, this approach relaxes the key constraint of linearity. While it does not lead to full equivalence to non-linear SFA, it recasts the classical slowness objective in a variational framework. Specifically, it allows the slowness objective to be interpreted as a regularizer to a reconstruction loss. Furthermore, we provide arguments, why -- from the perspective of slowness optimization -- the reconstruction loss takes on the role of the constraints that ensure informativeness in SFA. We conclude with a discussion of potential new research directions.
\end{abstract}

\section{Introduction}
\definecolor{cobalt}{rgb}{0.0, 0.28, 0.67}
\def \variationalcolor {cobalt}
\def \minnodesize {1cm}

Developing probabilistic perspectives on established machine learning algorithms can be a promising endeavor, as it casts methods originating from, for example, geometric or heuristic concepts into a well-understood framework that allows one to make explicit the assumptions and the dependencies that are inherent in the resulting model.
Many methods have been described in this shared language, even spanning the broad machine learning paradigms of unsupervised, supervised, and reinforcement learning. This makes it possible to compare methods, understand shortcomings, and propose extensions through a rich body of broad research.

Furthermore, previous research on a specific method that was generalized in such a way might prove to be useful for the field of probabilistic modeling itself. After all, the most efficient methods for probabilistic inference under a model are rarely the most general and often leverage the model-specific structure \parencite{Kalman1960, Margossian2024}.

In this work, a soft variant of Slow Feature Analysis (SFA) \parencite{Wiskott1998, WiskottSejnowski2002} is derived using the language of probabilistic inference. In SFA, a series of mappings $g_i$ from the samples to the low-dimensional representation learned so that they optimize
\begin{mini!}   
  {g_i}{\big<(g_i\left(\mathbf{x}_{t+1}) - g_i(\mathbf{x}_{t})\right)^2\big>_t}{}{}\label{eq:sfa_objective}
  \addConstraint{\big<g_i(\mathbf{x}_t)\big>_t=0}{}{}
  \addConstraint{\big<g_i(\mathbf{x}_t)g_j(\mathbf{x}_t)\big>_t=0,}{}{\quad\forall j < i}\label{eq:sfa_decorrelation_constraint}
  \addConstraint{\big<g_i(\mathbf{x}_t)^2\big>_t=1,}{}{\quad\forall i}\label{eq:sfa_unitvariance_constraint}
\end{mini!}
\noindent
where $\big< \cdot \big>_t$ is the average over time. The constraints fulfill different roles: The unit-variance constraint \eqref{eq:sfa_unitvariance_constraint} avoids the trivial solution of producing constant and thus uninformative features. The decorrelation constraint \eqref{eq:sfa_decorrelation_constraint} prohibits the learning of redundant mappings, so that, for $i\neq j$, $g_i$ and $g_j$ are different mappings. Solving this optimization problem leads to a set of mappings, ordered by their respective slowness. In this work, we discard this ordering for simplification, leading to the unordered objective:
\begin{align}
    \big<\|g(\mathbf{x}_{t+1}) - g(\mathbf{x}_{t})\|^2\big>_t \label{eq:sfa_unordered_objective}
\end{align}
where $g$ is vectorial.

As described the work presented, this is far from the first approach to view SFA probabilistically. However, in contrast to other approaches, we do find SFA not as an optimal solution to a probabilistic query, but as the result of severe limitations to the variational family used for approximating such an optimal solution. To that end, we alleviate assumptions on the actual model (specifically, linearity), but sacrifice exact equivalence to canonical (hard) SFA.

\section{Related Work}
Previous probabilistic perspectives on SFA as a latent variable model have been developed, and the topic has attracted renewed interest in the last decade. 
Almost all subsequent work in this direction, including the one presented, is based on an initial treatment of probabilistic SFA (later called PSFA) \parencite{Turner2007} in which SFA was derived from a linear Gaussian state space model \parencite{Roweis1999} with linear emissions. Linear SFA was recovered as the maximum-likelihood estimate of the emission model, with exact equivalence only in the limit of zero observation noise and annealing of the internal dynamics towards white noise \footnote{Although the latter mainly for scale correction.}.

Multiple extensions to this model have been proposed to adapt to the properties of real-world data, such as including non-zero observation noise \parencite{Omori2013}, increasing robustness to outliers through t-distributed observation noise \parencite{Fan2018}, modeling change points in the underlying dynamics \parencite{Omori2020}, and non-linearities \parencite{Puli2024}. 

The latter extension is most related to this work: It does not make strong assumptions regarding linearity and uses the general framework of variational inference. However, the presented approach differs substantially in its relation to SFA as well as in the assumed inference model. 

Another closely related area of research are variational autoencoders (VAE) \parencite{Kingma2014} and, in particular, dynamic variational autoencoders (DVAE) \parencite{Girin2021}, a conceptual umbrella for a number of models and variational inference approaches that share serial latent-space structure and conditional independence assumptions. In fact, it is the same structure assumed by PSFA. From this perspective, the presented approach can also be seen as DVAE.

The following section introduces the key concepts of variational inference, illustrating the general framework aiding in understanding the motivation and implications of the later results. 

\newpage
\section{Variational Inference}
\setlength{\belowcaptionskip}{-8pt}
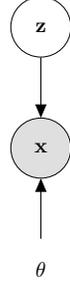
\begin{wrapfigure}{r}{0.15\textwidth}
\centering

\scalebox{0.8}{
\begin{tikzpicture}
  \node[obs, minimum size=\minnodesize]                               (x) {$\mathbf{x}$};
  \node[latent, above=of x, xshift=0cm, minimum size=\minnodesize] (z) {$\mathbf{z}$};
  \node[const, below=of x, inner sep=4pt, minimum size=\minnodesize] (theta) {$\theta$};
  \edge {z, theta} {x} ; %
\end{tikzpicture}
}
    \caption{Basic assumption in graph notation.}
    \label{fig:vaesfa_basic_model}
\end{wrapfigure}
\setlength{\belowcaptionskip}{0pt}
The setting underlying this research is based on the assumption that any observable data are sampled from a data-generating distribution $p(x)$ and that this distribution can be modeled by a parameterized distribution with non-trivial internal structure 
\begin{equation}
    \label{eq:svi_marginal}
    p(x) = \int p_\theta(x|z)p(z)dz
\end{equation} where $x$ are observable random variables, $z$ are latent random variables, and $\theta$ are model parameters to be identified. While the distribution $p_\theta(x|z)$ (here called the forward model) is typically complex and flexible, the prior distribution $p(z)$ is typically chosen to be a simple high-entropy distribution, such as a Gaussian. Figure \ref{fig:vaesfa_basic_model} shows the assumption in basic graphical model notation.

The objective is to find the maximum-likelihood estimate 
\begin{equation}
        \theta^* = {\text{arg}\max}_\theta \log p_\theta(\mathcal{X}) \label{eq:vaesfa_mle_objective}
\end{equation}
for a given dataset $\mathcal{X}$. In complex models, $p_\theta(x)$ has neither a closed form nor tractable estimators, if not extensively researched for a specific model, thus eluding direct optimization of $\theta$. 

A way to solve this is \textit{(stochastic) variational inference (VI)}, a methodology for approximate inference in probabilistic models. It has seen extensive research since its conception \parencite{Saul1996, Jordan1999, Hoffman13a, Ranganath14} with one of the prime examples of recent use in machine learning being the Variational Autoencoder \parencite{Kingma13_VAE}. VI heavily employs the decomposition
\begin{equation}
    \log p_\theta(x) = \underbrace{\mathbb{E}_{z\sim q_\phi}\left[\log\frac{p_\theta(x, z)}{q_\phi(z)}\right]}_{\mathcal{L}(x, \theta, \phi)} + KL\Big(q_\phi(z)\|p_\theta(z|x)\Big) \label{eq:svi_elbo}
\end{equation}
for a distribution $q_\phi(z)$ and $\mathcal{L}$. The former is known as the \textit{variational distribution}, the latter is known as the \textit{evidence lower-bound (ELBO)} and has multiple interesting properties. 
It is tractable and can be further decomposed into 
\begin{align}
    \mathcal{L}(x, \theta, \phi) &= \mathbb{E}_{z\sim q_\phi}\left[\log\frac{p_\theta(x, z)}{q_\phi(z)}\right]\\
    &=  \mathbb{E}_{z\sim q_\phi}\left[\log p_\theta(x|z) + \log p(z) - \log q_\phi(z) \vphantom{\frac{0}{0}}\right]\\
    &=  \mathbb{E}_{z\sim q_\phi}\left[\log p_\theta(x|z) - \log \frac{q_\phi(z)}{p(z)}\right]\\
    &=  \mathbb{E}_{z\sim q_\phi}\left[\log p_\theta(x|z)\vphantom{\frac{0}{0}}\right] - KL\Big(q_\phi(z)\|p(z)\Big). \label{eq:svi_elbo_decomp}
\end{align}
When viewed as a loss function, the left part can be understood as a reconstruction error\footnote{For example, in an autoencoder model.} while the Kulback-Leibler divergence takes the role of a regularizer on the variational distribution. Most importantly, since the KL divergence is non-negative, $\mathcal{L}$ is a lower bound on the typically intractable objective:
\begin{equation}
    \mathcal{L}(x, \theta, \phi) \leq \log p_\theta(x).
\end{equation}

Given a sufficiently flexible $q_\phi$, any effective optimization of $\mathcal{L}$ w.r.t. $\phi$ will inevitably tighten the bound and a subsequent optimization step w.r.t. $\theta$ will thus improve on the objective \eqref{eq:vaesfa_mle_objective}. As the bound becomes tight if and only if $q_\phi(z) = p_\theta(z|x)$, the variational distribution $q_\phi(z)$ is generally considered to be an approximate of the posterior $p_\theta(z|x)$.
In practice, both $\phi$ and $\theta$ often parameterize neural networks or other differentiable approximators and both are optimized in parallel through a variant of stochastic-gradient descent. In these cases, $\phi$ is shared for all data and $q_\phi$ is implemented as distribution conditioned on data,
\begin{equation}
    q_\phi(z) = q_\phi(z|x).
\end{equation} This setup is known as amortized inference \parencite{Gershman2014Amortized} and is well suited for larger datasets. 

\setlength{\belowcaptionskip}{-2pt}
\begin{figure}[h]
\centering
\scalebox{0.8}{
\begin{tikzpicture}
  \node[latent, xshift=0cm, minimum size=\minnodesize] (z1) {$\mathbf{z}_i$};
  \node[latent, right=of z1, minimum size=\minnodesize] (z2) {$\mathbf{z}_{i+1}$};
  \node[latent, left=of z1, minimum size=\minnodesize] (z0)  {$\mathbf{z}_{i-1}$};
  \node[obs, below=of z1, minimum size=\minnodesize] (x1) {$\mathbf{x}_i$};
  \node[obs, below=of z0, left=of x1, minimum size=\minnodesize] (x0) {$\mathbf{x}_{i-1}$};
  \node[obs, below=of z2, right=of x1, minimum size=\minnodesize] (x2) {$\mathbf{x}_{i+1}$};
  \node[const, below=of x1, inner sep=8pt, yshift=-10pt] (theta) {$\mathbf{\theta}$};
  \node[const, above=of z1, inner sep=8pt, yshift=+10pt, color=\variationalcolor] (phi) {$\phi$};
  \node[const, right=of z2, inner sep=8pt] (dotsright) {$\cdots$};
  \node[const, left=of z0, inner sep=8pt] (dotsleft) {$\cdots$};

  \node[const, right=of x2, inner sep=8pt] (dotsrightbottom) {$\cdots$};
  \node[const, left=of x0, inner sep=8pt] (dotsleftbottom) {$\cdots$};

  \edge {z1, theta} {x1} ; %
  \edge {z2, theta} {x2} ; %
  \edge {z0, theta} {x0} ; %
  \edge {theta} {dotsrightbottom, dotsleftbottom};

  \edge[dashed, \variationalcolor] {phi} {z1, z0, z2, dotsleft, dotsright} ;

    \path (x0) edge [->, >={triangle 45}, dashed, \variationalcolor, bend right=45] (z0);
  \path (x1) edge [->, >={triangle 45}, dashed, \variationalcolor, bend right=45] (z1);
  \path (x2) edge [->, >={triangle 45}, dashed, \variationalcolor, bend right=45] (z2);
\end{tikzpicture}
}
    \caption{Assumptions (generative and variational) in graphical model notation for multiple data points. Variational approximations indicated blue and with dashed lines.}
    \label{fig:vaesfa_basic_model_with_var}
\end{figure}
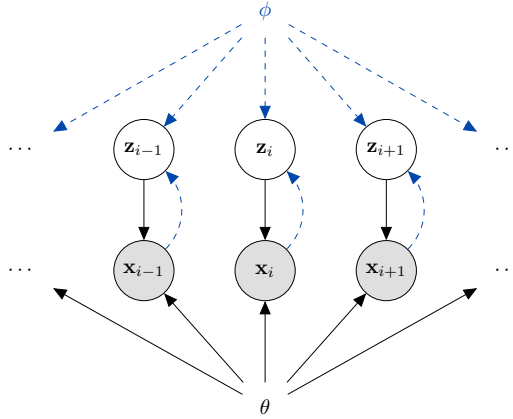
\setlength{\belowcaptionskip}{0pt}
Figure \ref{fig:vaesfa_basic_model_with_var} shows the graphical model notation of the generative assumptions, together with the assumptions on the variational distribution marked in blue and with dashed lines. The random variable, the posterior distribution of which is the target of the approximation, and the conditioning variables that are used in the approximation are implied by the target and the source of the arrows, respectively. This notation is used throughout the rest of this work.

\section{Generative Model}
\label{sec:sfavae_generative_model}
As SFA is concerned with time series data, it is assumed that the observable data are a process $\mathbf{x}_{1:T}$ of length $T$ consisting of real values $\mathbf{x}_t$. Correspondingly, we assume a real-valued latent process $\mathbf{z}_{1:T}$ of the same length.
This section defines the model in two parts as in eq. \ref{eq:svi_marginal} -- a prior distribution $p(\mathbf{z}_{1:T})$ and the forward model of observables $p(\mathbf{x}_{1:T}|\mathbf{z}_{1:T})$.

\paragraph{Prior} A prior distribution with Markov chain structure is assumed. It thus factorizes according to
\begin{align}
    p(\mathbf{z}_{1:T}) = p(\mathbf{z}_1)\prod_{t=2}^T p(\mathbf{z}_{t}|\mathbf{z}_{t-1}). \label{eq:sfavae_prior_assumption}
\end{align}
For each conditional in the chain, it is assumed that 
\begin{align}
    p(\mathbf{z}_{t}|\mathbf{z}_{t-1}) = \mathcal{N}(\mathbf{z}_{t-1}, \mathbf{I}) \label{eq:sfavae_prior}
\end{align}
thus resulting in a Gaussian state space model with independent dimensions, similar to PSFA's latent prior. In contrast to \textcite{Turner2007}, who assumed the next step to be centered on a linear transformation $\mathbf{\Lambda} z_{t-1}$ using a diagonal $\mathbf{\Lambda}$ with ordered diagonal elements $\lambda_i$, no meaningful transformation of the mean is assumed, i.e., $\lambda_i = 1, \forall i$. This is one of the most fundamental changes compared to PSFA: The slowness hypothesis is less clearly woven into the prior assumption and instead emerges from inference. It also implies that the prior process does not settle into a stationary distribution.
As an additional side-effect, ordering is disregarded in this model, but can in practice be established after feature extraction once a suitable subspace was identified.

\paragraph{Forward model} Interestingly, the presented argument does not require many assumptions on the forward model. The definition of the state-space model is completed by assuming that any observable $\mathbf{x}_t$ is only dependent on $\mathbf{z}_t$ and conditionally independent of all other variables. Thus,
\begin{align}
    p_\theta(\mathbf{x}_{1:T}|\mathbf{z}_{1:T}) = \prod_{t=1}^T p_\theta(\mathbf{x}_t|\mathbf{z}_{t}). \label{eq:sfavae_forward_model}
\end{align}
Another assumption not formalized is the differentiability w.r.t. $\theta$. Specifically, one could assume a neural network model that produces the statistics of a distribution that fits the observed data, such as the success probabilities of independent Bernoulli bits for image data or the mean and variance of a Gaussian for continuous process data, as done in variational autoencoders.
\setlength{\belowcaptionskip}{0pt}
\begin{figure}[h]
\centering
\scalebox{0.8}{
\begin{tikzpicture}
  \node[latent, xshift=0cm, minimum size=\minnodesize] (z1) {$\mathbf{z}_t$};
  \node[latent, right=of z1, minimum size=\minnodesize] (z2) {$\mathbf{z}_{t+1}$};
  \node[latent, left=of z1, minimum size=\minnodesize] (z0)  {$\mathbf{z}_{t-1}$};
  \node[obs, below=of z1, minimum size=\minnodesize] (x1) {$\mathbf{x}_t$};
  \node[obs, below=of z0, left=of x1, minimum size=\minnodesize] (x0) {$\mathbf{x}_{t-1}$};
  \node[obs, below=of z2, right=of x1, minimum size=\minnodesize] (x2) {$\mathbf{x}_{t+1}$};
  \node[const, below=of x1, inner sep=8pt, yshift=-10pt] (theta) {$\theta$};
  \node[const, right=of z2, inner sep=8pt] (dotsright) {$\cdots$};
  \node[const, left=of z0, inner sep=8pt] (dotsleft) {$\cdots$};

  \node[const, right=of x2, inner sep=8pt] (dotsrightbottom) {$\cdots$};
  \node[const, left=of x0, inner sep=8pt] (dotsleftbottom) {$\cdots$};

  \edge {z1, theta} {x1} ; %
  \edge {z2, theta} {x2} ; %
  \edge {z0, theta} {x0} ; %
  \edge {theta} {dotsrightbottom, dotsleftbottom};
  \edge {z1} {z2} ; %
  \edge {z0} {z1} ; %
  \edge {dotsleft} {z0} ; %
  \edge {z2} {dotsright} ; %
  
\end{tikzpicture}
}
    \caption{Generative assumptions with structured prior and forward model.}
    \label{fig:vaesfa_generative_process_model}
\end{figure}
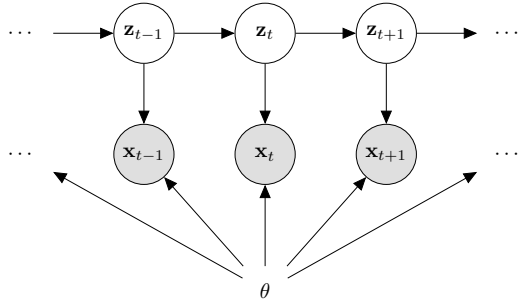
\setlength{\belowcaptionskip}{0pt}

The generative model is shown in Figure \ref{fig:vaesfa_generative_process_model} in graphical notation.

\section{Variational Model}
\label{sec:sfavae_variational_model}
A typical notion in variational inference is to choose $q_\phi$ to mirror the dependency structure (factorization) of the posterior to be approximated.
Note that many options are possible, depending on the scenario and algorithmic needs. 
A full latent posterior given all observables could be factorized according to the Markov structure of the model
\begin{align}
        p_\theta(\mathbf{z}_{1:T}|\mathbf{x}_{1:T}) = \prod_{t=1}^T p_\theta(\mathbf{z}_{t}|\mathbf{z}_{1:t-1}, \mathbf{x}_{1:T}) = \prod_{t=1}^T p_\theta(\mathbf{z}_{t}|\mathbf{z}_{t-1}\mathbf{x}_{t:T})
\end{align}
and a variational distribution $q_\phi$ could be chosen to factorize accordingly:
\begin{align}
            q_\phi(\mathbf{z}_{1:T}|\mathbf{x}_{1:T}) = \prod_{t=1}^T q_\phi(\mathbf{z}_{t}|\mathbf{z}_{1:t-1}, \mathbf{x}_{1:T}) = \prod_{t=1}^T q_\phi(\mathbf{z}_{t}|\mathbf{z}_{t-1}\mathbf{x}_{t:T}). \label{eq:sfvae_full_q_factorization}
\end{align}
However, the factors carry involved conditionals, in this case, requiring an estimate of the latent state as well as the future observations. 
Additional simplifications are very common, such as a mean-field assumption
\begin{align}
            q_\phi(\mathbf{z}_{1:T}|\mathbf{x}_{1:T}) = \prod_{t=1}^T q_\phi(\mathbf{z}_{t}|\mathbf{x}_{1:T}) \label{eq:sfavae_meanfield_q_factorization}
\end{align}
possibly combined with a causal flow of time to ensure applicability to streaming data:
\begin{align}
            q_\phi(\mathbf{z}_{1:T}|\mathbf{x}_{1:T}) = \prod_{t=1}^T q_\phi(\mathbf{z}_{t}|\mathbf{x}_{1:t}). \label{eq:sfavae_causal_q_factorization}
\end{align}
This work makes even more drastic simplifications by only allowing instantaneous factors as
\begin{align}
            q_\phi(\mathbf{z}_{1:T}|\mathbf{x}_{1:T}) = \prod_{t=1}^T q_\phi(\mathbf{z}_{t}|\mathbf{x}_{t}). \label{eq:sfavae_variational_factorization}
\end{align}
This already hints at the role of the variational distribution when investigating the relationship to SFA, which learns instantaneous feature extractors that respect time-structure in observed data by applying the same mapping to all data points individually. The factors in equation \ref{eq:sfavae_variational_factorization} mimic this mode of application.

\def \scalingfactor {0.53}

\begin{figure}[t]
    \centering
    \begin{subfigure}[b]{0.3\textwidth}
        \centering
        \scalebox{\scalingfactor} {
            \begin{tikzpicture}
              \node[latent, xshift=0cm, minimum size=\minnodesize, text=\variationalcolor, draw=\variationalcolor] (z1) {$\mathbf{z}_t$};
              \node[latent, right=of z1, minimum size=\minnodesize] (z2) {$\mathbf{z}_{t+1}$};
              \node[latent, left=of z1, minimum size=\minnodesize] (z0)  {$\mathbf{z}_{t-1}$};
              \node[obs, below=of z1, minimum size=\minnodesize] (x1) {$\mathbf{x}_t$};
              \node[obs, below=of z0, left=of x1, minimum size=\minnodesize] (x0) {$\mathbf{x}_{t-1}$};
              \node[obs, below=of z2, right=of x1, minimum size=\minnodesize] (x2) {$\mathbf{x}_{t+1}$};
              \node[const, below=of x1, inner sep=8pt, yshift=-10pt] (theta) {$\theta$};
              \node[const, above=of z1, inner sep=8pt, yshift=+10pt, color=\variationalcolor] (phi) {$\phi$};
              \node[const, right=of z2, inner sep=8pt] (dotsright) {$\cdots$};
              \node[const, left=of z0, inner sep=8pt] (dotsleft) {$\cdots$};

              \node[const, right=of x2, inner sep=8pt] (dotsrightbottom) {$\cdots$};
              \node[const, left=of x0, inner sep=8pt] (dotsleftbottom) {$\cdots$};
            
              \edge {z1, theta} {x1} ; %
              \edge {z2, theta} {x2} ; %
              \edge {z0, theta} {x0} ; %
              \edge {theta} {dotsrightbottom, dotsleftbottom};
              \edge {z1} {z2} ; %
              \edge {z0} {z1} ; %
              \edge {dotsleft} {z0} ; %
              \edge {z2} {dotsright} ; %
            
              \edge[dashed, \variationalcolor] {phi} {z1} ;
              
              \path (z0) edge [->, >={triangle 45}, dashed, \variationalcolor, bend left=30] (z1);

              \path (x1) edge [->, >={triangle 45}, dashed, \variationalcolor, bend right=20] (z1);
              \path (x2) edge [->, >={triangle 45}, dashed, \variationalcolor, ] (z1);
              
              \path (dotsrightbottom) edge [->, >={triangle 45}, dashed, \variationalcolor] (z1);
            \end{tikzpicture}

            }
        \caption{Posterior-like factorization}
        \label{fig:first}
    \end{subfigure}
    \hfill
    \begin{subfigure}[b]{0.3\textwidth}
        \centering
        \scalebox{\scalingfactor} {
            \begin{tikzpicture}
              \node[latent, xshift=0cm, minimum size=\minnodesize, text=\variationalcolor, draw=\variationalcolor] (z1) {$\mathbf{z}_t$};
              \node[latent, right=of z1, minimum size=\minnodesize] (z2) {$\mathbf{z}_{t+1}$};
              \node[latent, left=of z1, minimum size=\minnodesize] (z0)  {$\mathbf{z}_{t-1}$};
              \node[obs, below=of z1, minimum size=\minnodesize] (x1) {$\mathbf{x}_t$};
              \node[obs, below=of z0, left=of x1, minimum size=\minnodesize] (x0) {$\mathbf{x}_{t-1}$};
              \node[obs, below=of z2, right=of x1, minimum size=\minnodesize] (x2) {$\mathbf{x}_{t+1}$};
              \node[const, below=of x1, inner sep=8pt, yshift=-10pt] (theta) {$\theta$};
              \node[const, above=of z1, inner sep=8pt, yshift=+10pt, color=\variationalcolor] (phi) {$\phi$};
              \node[const, right=of z2, inner sep=8pt] (dotsright) {$\cdots$};
              \node[const, left=of z0, inner sep=8pt] (dotsleft) {$\cdots$};

              \node[const, right=of x2, inner sep=8pt] (dotsrightbottom) {$\cdots$};
              \node[const, left=of x0, inner sep=8pt] (dotsleftbottom) {$\cdots$};
            
              \edge {z1, theta} {x1} ; %
              \edge {z2, theta} {x2} ; %
              \edge {z0, theta} {x0} ; %
              \edge {theta} {dotsrightbottom, dotsleftbottom};
              \edge {z1} {z2} ; %
              \edge {z0} {z1} ; %
              \edge {dotsleft} {z0} ; %
              \edge {z2} {dotsright} ; %
            
              \edge[dashed, \variationalcolor] {phi} {z1} ;

              \path (x1) edge [->, >={triangle 45}, dashed, \variationalcolor, bend right=20] (z1);
              \path (x2) edge [->, >={triangle 45}, dashed, \variationalcolor, ] (z1);
              \path (x0) edge [->, >={triangle 45}, dashed, \variationalcolor, bend right=0] (z1);
              
              \path (dotsrightbottom) edge [->, >={triangle 45}, dashed, \variationalcolor] (z1);
              \path (dotsleftbottom) edge [->, >={triangle 45}, dashed, \variationalcolor] (z1);
            \end{tikzpicture}
            }
        \caption{Mean-field assumption}
        \label{fig:second}
    \end{subfigure}
        \hfill
    \begin{subfigure}[b]{0.3\textwidth}
        \centering
        \scalebox{\scalingfactor} {
            \begin{tikzpicture}
              \node[latent, xshift=0cm, minimum size=\minnodesize, text=\variationalcolor, draw=\variationalcolor] (z1) {$\mathbf{z}_t$};
              \node[latent, right=of z1, minimum size=\minnodesize] (z2) {$\mathbf{z}_{t+1}$};
              \node[latent, left=of z1, minimum size=\minnodesize] (z0)  {$\mathbf{z}_{t-1}$};
              \node[obs, below=of z1, minimum size=\minnodesize] (x1) {$\mathbf{x}_t$};
              \node[obs, below=of z0, left=of x1, minimum size=\minnodesize] (x0) {$\mathbf{x}_{t-1}$};
              \node[obs, below=of z2, right=of x1, minimum size=\minnodesize] (x2) {$\mathbf{x}_{t+1}$};
              \node[const, below=of x1, inner sep=8pt, yshift=-10pt] (theta) {$\mathbf{\theta}$};
              \node[const, above=of z1, inner sep=8pt, yshift=+10pt, color=\variationalcolor] (phi) {$\phi$};
              \node[const, right=of z2, inner sep=8pt] (dotsright) {$\cdots$};
              \node[const, left=of z0, inner sep=8pt] (dotsleft) {$\cdots$};

              \node[const, right=of x2, inner sep=8pt] (dotsrightbottom) {$\cdots$};
              \node[const, left=of x0, inner sep=8pt] (dotsleftbottom) {$\cdots$};
            
              \edge {z1, theta} {x1} ; %
              \edge {z2, theta} {x2} ; %
              \edge {z0, theta} {x0} ; %
              \edge {theta} {dotsrightbottom, dotsleftbottom};
              \edge {z1} {z2} ; %
              \edge {z0} {z1} ; %
              \edge {dotsleft} {z0} ; %
              \edge {z2} {dotsright} ; %
            
              \edge[dashed, \variationalcolor] {phi} {z1} ;
              
              \path (x1) edge [->, >={triangle 45}, dashed, \variationalcolor, bend right=20] (z1);
            
            \end{tikzpicture}
            }
        \caption{Instantaneous factorization}
        \label{subfig:sfavae_factorizations_instant}
    \end{subfigure}
    \caption{Comparing three types of factors for the variational distribution. This work chooses the factorization in Subfigure \ref{subfig:sfavae_factorizations_instant}.}
    \label{fig:sidebyside}
\end{figure}
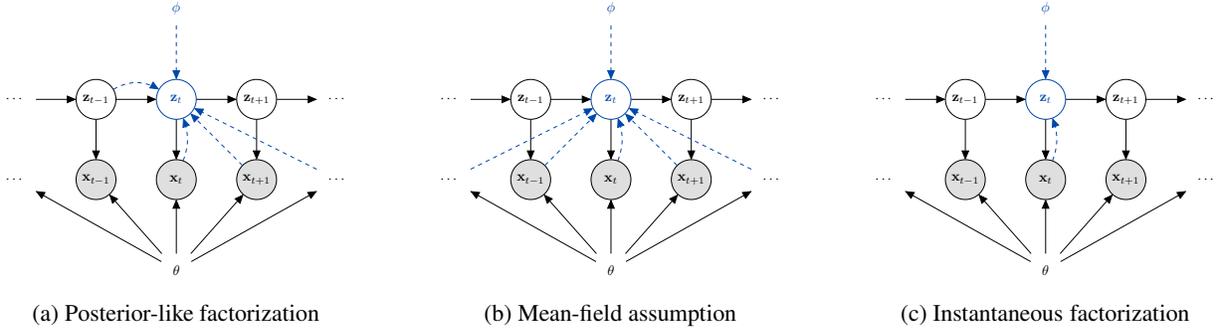

As opposed to the forward model, this work makes more detailed design decisions on the nature of these factors:
\begin{align}
    q_\phi(\mathbf{z}_t|\mathbf{x}_t) = \mathcal{N}(\mathbf{z}_t| \mathbf{g}_\phi(\mathbf{x}_t), \mathbf{I}) \label{eq:sfavae_variational_single_factor}
\end{align}
where $\mathbf{g}_\phi$ is a differentiable approximator, such as a neural network.

\section{Combining Both Models}
\label{sec:vaesfa_elbo-derivation}
In the following section, the implications of combining the generative assumptions discussed in Section \ref{sec:sfavae_generative_model} with the design of the variational distribution discussed in Section \ref{sec:sfavae_variational_model} for variational inference are derived. Specifically, it is shown how the ELBO objective (eq. \eqref{eq:svi_elbo}) can be understood as a variant of the SFA optimization problem.
For process data, the ELBO can be defined as 
\begin{align}
    \mathcal{L}(\mathbf{x}_{1:T}, \theta, \phi) &= \mathbb{E}_{\mathbf{z}_{1:T}\sim q_\phi}\left[\log\frac{p_\theta(\mathbf{x}_{1:T}, \mathbf{z}_{1:T})}{q_\phi(\mathbf{z}_{1:T}|\mathbf{x}_{1:T})}\right] 
    \intertext{and following the decomposition \eqref{eq:svi_elbo_decomp} as}
    &= \mathbb{E}_{\mathbf{z}_{1:T}\sim q_\phi} \left[\log p_\theta(\mathbf{x}_{1:T} | \mathbf{z}_{1:T}) \vphantom{\frac{0}{0}}\right] - KL\Big(q_\phi(\mathbf{z}_{1:T}|\mathbf{x}_{1:T})\|p(\mathbf{z}_{1:T})\vphantom{\frac{0}{0}}\Big).
\end{align}
In the following, both parts of the ELBO are simplified individually using the structural assumptions made previously.
\paragraph{Left side} 
Reducing the left side to a simpler form is straightforward and relies only on the assumption of the forward model and the properties of the expected value operator.
\begin{align}
    \mathbb{E}_{\mathbf{z}_{1:T}\sim q_\phi} \left[\log p_\theta(\mathbf{x}_{1:T} | \mathbf{z}_{1:T}) \vphantom{\frac{0}{0}}\right] &\stackrel{\eqref{eq:sfavae_forward_model}}{=} \mathbb{E}_{\mathbf{z}_{1:T}\sim q_\phi} \left[\sum_t \log p_\theta(\mathbf{x}_{t} | \mathbf{z}_{t}) \vphantom{\frac{0}{0}}\right]\\
    &\stackrel{\hphantom{\eqref{eq:sfavae_forward_model}}}{=} \sum_t \mathbb{E}_{\mathbf{z}_{1:T}\sim q_\phi} \left[\log p_\theta(\mathbf{x}_{t} | \mathbf{z}_{t}) \vphantom{\frac{0}{0}}\right] \\
    &\stackrel{\hphantom{\eqref{eq:sfavae_forward_model}}}{=} \sum_t \mathbb{E}_{\mathbf{z}_{t}\sim q_\phi} \left[\log p_\theta(\mathbf{x}_{t} | \mathbf{z}_{t}) \vphantom{\frac{0}{0}}\right] \label{eq:elbo_rec_part}
\end{align}
Thus, the reconstruction loss becomes a sum of reconstruction losses of individual data points, similar to a batch reconstruction loss for (variational) autoencoders.

\paragraph{Right side} Reducing the right side is more involved, but apart from the referenced assumptions it is based mainly on simple operations. The derivation is included in some detail as clarifies the influence of assumptions of the covariance structure of the different parts of the model. This leads to what the main insight of this work is, and can aid possible future extensions.
\begin{align}
&KL\Big(q_\phi(\mathbf{z}_{1:T}|\mathbf{x}_{1:T})\|p(\mathbf{z}_{1:T})\vphantom{\frac{0}{0}}\Big)\\
    &\quad\quad\quad\stackrel{\text{def.}}{=} \mathbb{E}_{\mathbf{z}_{1:T}\sim q_\phi} \left[\log q_\phi(\mathbf{z}_{1:T}|\mathbf{x}_{1:T}) - \log p(\mathbf{z}_{1:T})\vphantom{\frac{0}{0}}\right]\\
     &\quad\quad\quad\stackrel{\eqref{eq:sfavae_variational_factorization}}{=} \mathbb{E}_{\mathbf{z}_{1:T}\sim q_\phi} \left[\sum_t \log q_\phi(\mathbf{z}_{t}|\mathbf{x}_{t}) - \log p(\mathbf{z}_{1:T})\vphantom{\frac{0}{0}}\right]\\
     &\quad\quad\quad\stackrel{\eqref{eq:sfavae_prior_assumption}}{=} \mathbb{E}_{\mathbf{z}_{1:T}\sim q_\phi} \left[\sum_t \log q_\phi(\mathbf{z}_{t}|\mathbf{x}_{t}) - \sum_t \log p(\mathbf{z}_{t}|\mathbf{z}_{t-1})\vphantom{\frac{0}{0}}\right]\\
     &\quad\quad\quad\stackrel{\hphantom{\text{def.}}}{=} \sum_t \mathbb{E}_{\mathbf{z}_{1:T}\sim q_\phi} \left[\log q_\phi(\mathbf{z}_{t}|\mathbf{x}_{t}) - \log p(\mathbf{z}_{t}|\mathbf{z}_{t-1})\vphantom{\frac{0}{0}}\right]\\
     &\quad\quad\quad\stackrel{\hphantom{\text{def.}}}{=} \sum_t \mathbb{E}_{\mathbf{z}_{t-1},\mathbf{z}_{t}\sim q_\phi} \left[\log q_\phi(\mathbf{z}_{t}|\mathbf{x}_{t}) - \log p(\mathbf{z}_{t}|\mathbf{z}_{t-1})\vphantom{\frac{0}{0}}\right]\\
     &\quad\quad\quad\stackrel{\hphantom{\text{def.}}}{=} \sum_t \mathbb{E}_{\mathbf{z}_{t-1}\sim q_\phi} \left[ \mathbb{E}_{\mathbf{z}_{t}\sim q_\phi} \left[\log q_\phi(\mathbf{z}_{t}|\mathbf{x}_{t}) - \log p(\mathbf{z}_{t}|\mathbf{z}_{t-1})\vphantom{\frac{0}{0}}\right]\right]\\
     &\quad\quad\quad\stackrel{\hphantom{\text{def.}}}{=} \sum_t \mathbb{E}_{\mathbf{z}_{t-1}\sim q_\phi}\left[KL\Big(q_\phi(\mathbf{z}_{t}|\mathbf{x}_{t})\|p(\mathbf{z}_{t}|\mathbf{z}_{t-1})\vphantom{\frac{0}{0}}\Big)\right]
     \intertext{At this point, the connection to SFA becomes increasingly apparent in the KL divergence between two subsequent points. Since both distributions are assumed and chosen to be Gaussian, the KL divergence has a convenient form, determined mostly by the difference in their means. Let $\mu_q$ and $\mu_p$ be the respective means of $q_\phi(\mathbf{z}_t|\mathbf{x}_t)$ and $p(\mathbf{z}_t|\mathbf{z}_{t-1})$ and $\Sigma_q$ and $\Sigma_p$ their covariance matrices. Then:}
     &\quad\quad\quad\stackrel{\hphantom{\text{def.}}}{=} \sum_t \mathbb{E}_{\mathbf{z}_{t-1}\sim q_\phi}\left[
        \frac{1}{2} \left(
            \|\mu_p - \mu_q\|^2_{\Sigma^{-1}_p} + \Tr(\Sigma^{-1}_p\Sigma_q)-\log\frac{|\Sigma_q|}{|\Sigma_p|} - d)
        \right)
     \right]
     \intertext{where $d$ is the dimension of $\mathbf{z}_t$. Since the covariances where chosen to $\Sigma_p = \Sigma_q = \mathbf{I}$, the term simplifies significantly. However, it should be noted that for any diagonal $\Sigma_p$ and $\Sigma_q$, only the $\Sigma_p^{-1}$-norm would change, corresponding to reweighting the contribution of individual features to the overall slowness loss. This becomes apparent after the following steps.}
     &\quad\quad\quad\stackrel{\hphantom{\eqref{eq:sfavae_prior}, \eqref{eq:sfavae_variational_single_factor}}}{=} \frac{1}{2} \sum_t \mathbb{E}_{\mathbf{z}_{t-1}\sim q_\phi}\left[ 
            \|\mu_p - \mu_q\|^2 \vphantom{\frac{0}{0}} - d
     \right]\\
     \intertext{From \eqref{eq:sfavae_variational_single_factor} follows $\mu_q=g_\phi(\mathbf{x}_t)$ and from \eqref{eq:sfavae_prior} follows $\mu_p = \mathbf{z}_{t-1}$. Note that the expectation is taken over $\mathbf{z}_{t-1}\sim\mathcal{N}(\mathbf{g}_\phi(\mathbf{x}_{t-1}),\mathbf{I})$. Simple manipulation leads to}
          &\quad\quad\quad\stackrel{\hphantom{\eqref{eq:sfavae_prior}, \eqref{eq:sfavae_variational_single_factor}}}{=} \frac{1}{2} \sum_t 
            \|\mathbf{g}_\phi(\mathbf{x}_{t-1}) - \mathbf{g}_\phi(\mathbf{x}_{t})\|^2 + d - d\\
          &\quad\quad\quad\stackrel{\hphantom{\eqref{eq:sfavae_prior}, \eqref{eq:sfavae_variational_single_factor}}}{=} \frac{1}{2} \sum_t 
            \|\mathbf{g}_\phi(\mathbf{x}_{t-1}) - \mathbf{g}_\phi(\mathbf{x}_{t})\|^2 \label{eq:elbo_slowness_part}
\end{align}
which is recognizable as an (unordered) slowness objective \eqref{eq:sfa_unordered_objective}.

\paragraph{Full ELBO}
Putting both sides together, the ELBO results in
\begin{align}
    &\mathcal{L}(\mathbf{x}_{1:T}, \theta, \phi) \\ &= \mathbb{E}_{\mathbf{z}_{1:T}\sim q_\phi} \left[\log p_\theta(\mathbf{x}_{1:T} | \mathbf{z}_{1:T}) \vphantom{\frac{0}{0}}\right] - KL\Big(q_\phi(\mathbf{z}_{1:T}|\mathbf{x}_{1:T})\|p(\mathbf{z}_{1:T})\vphantom{\frac{0}{0}}\Big)\\
    &=\sum_t \mathbb{E}_{\mathbf{z}_{t}\sim q_\phi} \left[\log p_\theta(\mathbf{x}_{t} | \mathbf{z}_{t}) \vphantom{\frac{0}{0}}\right] - \frac{1}{2} \sum_t 
            \|\mathbf{g}_\phi(\mathbf{x}_{t-1}) - \mathbf{g}_\phi(\mathbf{x}_{t})\|^2. \label{eq:sfavae_final_sfa_elbo}
\end{align}
Thus, the variational inference objective given all model assumptions would be equivalent to the parallel maximization of the reconstruction fidelity per point and minimization of slowness for the time-series.

\section{Interpretation and Perspectives}
\label{sec:sfavae_interpretation}
In this section, the implications and interpretations of the derived loss \eqref{eq:sfavae_final_sfa_elbo} are discussed. 

Equation \eqref{eq:elbo_slowness_part} is a slowness objective as previously defined, but the objective is not sufficient to elicit useful features as, for flexible enough $\mathbf{g}_\phi$, trivial solutions exist: Unconstrained optimization leads to constant features.

For this reason, the SFA optimization problem \ref{eq:sfa_objective} is formulated with unit-variance constraints to avoid constant features and zero-covariance constraints to avoid redundant features. Another variant of SFA \parencite{Bengio2009} includes these constraints as soft penalty for the same reason. 

Although not generally equivalent\footnote{For example, under reordering of the points, the relative weighting between reconstruction and slowness change in such a loss thereby putting more or less optimization pressure on either.}, we argue that reconstruction loss \eqref{eq:elbo_rec_part} plays a role comparable to both constraints, each through a different mechanism:
If the latent dimensionality is smaller than the input dimensionality, $g_\phi(\mathbf{z}|\mathbf{x})$ and the forward model $p_\theta(\mathbf{x}|\mathbf{z})$ can be understood as an encoder-decoder architecture, like the variational autoencoder or other autoencoders. In this case, the reduced dimensionality of the latent space acts as an information bottleneck with limited capacity. Due to this bottleneck, efficient compression is required for optimal reconstruction, naturally discouraging any redundancy in the latent representation in the form of correlation, thus fulfilling the role of the covariance constraint. For example, an optimal linear autoencoder recovers the same subspace as principal-component analysis \parencite{Baldi1989}, which is strongly related to SFA.

The correspondence to the unit-variance constraint is more subtle and is founded in the probabilistic formulation of the reconstruction loss. 
For illustration purposes, consider the reconstruction loss to be expressed as mean squared error $\|\mathbf{x}-\tilde{\mathbf{x}}\|^2$, where $\tilde{\mathbf{x}}$ is reconstruction after applying only the deterministic parts of encoder and decoder. In that case, note that for a sufficiently flexible architecture, any solution's slowness could be improved by scaling down the features in the encoder and reversing that scaling in the decoder without negatively impacting the reconstruction loss. The overall objective would thus not possess a finite optimizer and an effective optimization scheme would eventually produce numerical problems in practice. This is different for the probabilistic formulation
\begin{align*}
    \mathbb{E}_{\mathbf{z}\sim q_\phi} \Big[\log p_\theta(\mathbf{x} | \mathbf{z})\Big]
\end{align*}
as it assumes any $\mathbf{z}$ to be sampled according to $\mathcal{N}(\mathbf{z}|\mathbf{g}_\phi(\mathbf{x}), \mathbf{I})$ before decoding. Thus, for two different $\mathbf{x}$ and $\mathbf{x}'$  to be sufficiently distinguishable for reconstruction, the variation after encoding through $\mathbf{g}_\phi$ must be sufficiently large. Otherwise, the distributions of $\mathbf{z}$ and $\mathbf{z}'$ overlap, leading to poor expected reconstruction.
In that sense, the slowness loss and reconstruction loss are opposing objectives prohibiting indefinite up- or down-scaling, respectively, when the probabilistic formulation of the reconstruction loss is used. 

Since the probabilistic version of the loss requires evaluating the expectation, which does not generally have a closed form\footnote{Except for linear $\mathbf{g}_\phi$.}, sampling from $q_\phi(\mathbf{z}|\mathbf{x})$ is required for each individual data point. This is computationally expensive, as it has to be done anew for each evaluation of the loss, scales unfavorably with the latent space dimension, and induces a source of optimization noise, which is one reason why the reconstruction loss is generally replaced by the mean-squared error and reparameterization in VAEs.

Another interesting effect results from the typical perspectives taken on VAEs and SFA, respectively. While in VAEs, the KL divergence is typically seen as a regularizer to prevent the encoder network $\mathbf{g}_\phi$ from overfitting the reconstruction objective, the parts of the ELBO switch roles when seen from the SFA perspective:

\begin{align*}
        \mathcal{L}(\mathbf{x}_{1:T}, \theta, \phi) &= \overbrace{\mathbb{E}_{\mathbf{z}_{1:T}\sim q_\phi} \left[\log p_\theta(\mathbf{x}_{1:T} | \mathbf{z}_{1:T}) \vphantom{\frac{0}{0}}\right]}^{\text{VAE objective}} - \overbrace{KL\Big(q_\phi(\mathbf{z}_{1:T}|\mathbf{x}_{1:T})\|p(\mathbf{z}_{1:T})\vphantom{\left[\frac{0}{0}\right]}\Big)}^{\text{VAE regularizer}}\\
    &=\underbrace{\sum_t \mathbb{E}_{\mathbf{z}_{t}\sim q_\phi} \left[\log p_\theta(\mathbf{x}_{t} | \mathbf{z}_{t}) \vphantom{\frac{0}{0}}\right]}_{\text{SFA regularizer}} - \underbrace{\frac{1}{2} \sum_t 
            \|\mathbf{g}_\phi(\mathbf{x}_{t-1}) - \mathbf{g}_\phi(\mathbf{x}_{t})\|^2}_{\text{SFA objective}}.
\end{align*}
What was previously the regularizer becomes the main objective, while the previous objective becomes the regularizer to prevent trivial solutions. While this observation is not of immediate consequence, it might aid in the design and understanding of future extensions to SFA. One such possible extension might be an investigation paralleling the $\beta$-VAE \parencite{Higgins2017}, for which stronger regularization (by weighting the KL divergence with a scalar $\beta>1$ can lead to a more disentangled latent space, as well as employing insights from the same work regarding the relationship between architecture and regularization strength.

\paragraph{Linear Encoder and Decoder}
A more specific and non-deterministic formulation of the ELBO can be found in the case where both, encoder and decoder, are linear and Gaussian, i.e., where $q(\mathbf{z}|\mathbf{x}) = \mathcal{N}(\mathbf{z} |\mathbf{W}^T\mathbf{x} + \mathbf{b}, \mathbf{I})$ and $p(\mathbf{x}|\mathbf{z}) = \mathcal{N}(\mathbf{x} |\mathbf{V}\mathbf{z} + \mathbf{o}, \mathbf{I})$.

It is straightforward to confirm that the objective then becomes:
\begin{align}
    &\mathcal{L}(\mathbf{x}_{1:T}, \mathbf{W}^T, \mathbf{V}, \mathbf{o}, \mathbf{b})\nonumber\\&= \text{c} - \frac{T}{2}\Tr(\mathbf{V}^T\mathbf{V}) + \frac{1}{2}\sum_t \|\mathbf{x_t} - \tilde{\mathbf{x}}_t\|^2  - \frac{1}{2}\sum_t \|\mathbf{W}^T\mathbf{x}_{t-1} - \mathbf{W}^T\mathbf{x}_t\|^2
\end{align}
where $\tilde{\mathbf{x}}_t = \big(\mathbf{V}\mathbf{W}^T\mathbf{x}_t+\mathbf{Vb} + \mathbf{o}\big)$ is the deterministic reconstruction of $\mathbf{x}$ and $c$ is constant w.r.t. to the parameters and data. The term $\Tr(\mathbf{V}^T\mathbf{V})$ corresponds to a L2 regularization on the decoder weights and is the deterministic instantiation of the variance-preservation argument made in Section \ref{sec:sfavae_interpretation}.

Additionally, the simple form of the objective allows for determining necessary conditions for optimality.
From $\frac{\partial \mathcal{L}}{\partial \mathbf{V}}=\frac{\partial \mathcal{L}}{\partial \mathbf{W}^T}=\mathbf{0}$, $\frac{\partial \mathcal{L}}{\partial \mathbf{b}}=\mathbf{0}$ and $\frac{\partial \mathcal{L}}{\partial \mathbf{o}}=\mathbf{0}$ are equivalent to the two following conditions\footnote{We conveniently overload $\mathbf{0}$ to have dimensionality based on context.}:
\begin{align}
    \mathbf{W}^T\dot{\mathbf{C}}\mathbf{W} = \mathbf{V}^T\mathbf{V} \quad\text{and}\quad
    \mathbf{o} = \bar{\mathbf{x}} - \mathbf{VW}^T\bar{\mathbf{x}}-\mathbf{Vb} \label{eq:sfvae_necessary_cond_linear}
\end{align}
where $\bar{\mathbf{x}}$ is the time-series' mean and $ \dot{\mathbf{C}}$ is the second-moment matrix of step-wise differences. The derivation of equation \ref{eq:sfvae_necessary_cond_linear} can be found in the Appendix \ref{appendix:necessary_conditions_for_linear_sfavae}.

\section{Directions}
Connecting SFA to the active field of variational inference could yield benefits for both fields. In this section, we go through some research directions which we perceive as promising although not covered in this work:
\begin{description}
    \item[Variants of the forward model] The forward model has not been defined in detail for this investigation, but a wide range of variants is possible -- in practice, architectures are chosen specific to the data domain. Continuous sensor data might require a Gaussian parameterized by a neural network or a heavy-tailed distribution if prone to outliers, while for image data, independent Bernoulli distributions are common. The flexibility comes at the cost of increased burden in architectural design, but it allows for the inclusion of expert knowledge.
    \item[Generative properties] SFA is generally understood as dimensionality reduction method with no general means to reverse this reduction. The  perspective developed in this works yields the opportunity (and need) of co-training a generative structure, which can aid the interpretation of the extracted features, e.g., by means of latent space exploration. 
    \item[Extensions to the variational factorization] Similarly to the forward model, the variational model can be adapted to the data domain. Furthermore, it can be adapted to application scenario: For example, by using modern architectures like LSTMs \parencite{Hochreiter1997} or Transformer networks \parencite{Vaswani2017}, one could condition $q_\phi$ on all observables (corresponding to \eqref{eq:sfavae_meanfield_q_factorization}), if online application is not needed, or on an extended interval of observables, similar to the well-known Kalman smoothing \parencite{Kalman1960}. Conditioning on all past observables (corresponding to \eqref{eq:sfavae_causal_q_factorization}) would allow for accumulated state information and a way to deal with partial observability, which has not yet been extensively researched for SFA.  
    The approaches discussed by \textcite{Schueler2019} allow for similar modifications in an ad-hoc fashion, but the perspective of variational inference allows for such modifications in a principled way.
    \item[Inclusion of architectural whitening] The modifications by \textcite{Schueler2019} to allow for neural network training for SFA and the perspective developed in this work are not mutually exclusive: Architectural whitening can also be used in $\mathbf{g}_\phi$, albeit the interplay between the variational distribution and a whitened latent space is not trivially clear.
    \item[SFA as data-efficient initialization] As VI in general is computationally expensive, we hypothesize that variants of SFA for which good training heuristics exist, such as HSFA \parencite{Escalante2013}, could be used for data-efficient pre-training for initialization of $\mathbf{g}_\phi$, possibly transferring to other, related models. 
\end{description}

\printbibliography

@article{Hochreiter1997,
author = {Hochreiter, Sepp and Schmidhuber, J\"{u}rgen},
title = {Long Short-Term Memory},
year = {1997},
issue_date = {November 15, 1997},
publisher = {MIT Press},
address = {Cambridge, MA, USA},
volume = {9},
number = {8},
issn = {0899-7667},
url = {https://doi.org/10.1162/neco.1997.9.8.1735},
doi = {10.1162/neco.1997.9.8.1735},
journal = {Neural Comput.},
month = nov,
pages = {1735–1780},
numpages = {46}
}

@inproceedings{Vaswani2017,
author = {Vaswani, Ashish and Shazeer, Noam and Parmar, Niki and Uszkoreit, Jakob and Jones, Llion and Gomez, Aidan N. and Kaiser, \L{}ukasz and Polosukhin, Illia},
title = {Attention is all you need},
year = {2017},
isbn = {9781510860964},
publisher = {Curran Associates Inc.},
address = {Red Hook, NY, USA},
booktitle = {Proceedings of the 31st International Conference on Neural Information Processing Systems},
pages = {6000–6010},
numpages = {11},
location = {Long Beach, California, USA},
series = {NIPS'17}
}

@article{Kalman1960,
    author = {Kalman, R. E.},
    title = "{A New Approach to Linear Filtering and Prediction Problems}",
    journal = {Journal of Basic Engineering},
    volume = {82},
    number = {1},
    pages = {35-45},
    year = {1960},
    month = {03},
    issn = {0021-9223},
    doi = {10.1115/1.3662552},
    url = {https://doi.org/10.1115/1.3662552},
    eprint = {https://asmedigitalcollection.asme.org/fluidsengineering/article-pdf/82/1/35/5518977/35\_1.pdf},
}

@inproceedings{Higgins2017,
  author       = {Irina Higgins and
                  Lo{\"{\i}}c Matthey and
                  Arka Pal and
                  Christopher P. Burgess and
                  Xavier Glorot and
                  Matthew M. Botvinick and
                  Shakir Mohamed and
                  Alexander Lerchner},
  title        = {beta-VAE: Learning Basic Visual Concepts with a Constrained Variational
                  Framework},
  booktitle    = {5th International Conference on Learning Representations, {ICLR} 2017,
                  Toulon, France, April 24-26, 2017, Conference Track Proceedings},
  publisher    = {OpenReview.net},
  year         = {2017},
  url          = {https://openreview.net/forum?id=Sy2fzU9gl},
  bibsource    = {dblp computer science bibliography, https://dblp.org}
}

@article{Baldi1989,
title = {Neural networks and principal component analysis: Learning from examples without local minima},
journal = {Neural Networks},
volume = {2},
number = {1},
pages = {53-58},
year = {1989},
issn = {0893-6080},
doi = {https://doi.org/10.1016/0893-6080(89)90014-2},
url = {https://www.sciencedirect.com/science/article/pii/0893608089900142},
author = {Pierre Baldi and Kurt Hornik}
}

@inproceedings{
Margossian2024,
title={Amortized Variational Inference: When and Why?},
author={Charles Margossian and David Blei},
booktitle={The 40th Conference on Uncertainty in Artificial Intelligence},
year={2024},
url={https://openreview.net/forum?id=mCVYIsnctr}
}

@article{Saul1996,
author = {Saul, Lawrence K. and Jaakkola, Tommi and Jordan, Michael I.},
title = {Mean field theory for sigmoid belief networks},
year = {1996},
issue_date = {January 1996},
publisher = {AI Access Foundation},
address = {El Segundo, CA, USA},
volume = {4},
number = {1},
issn = {1076-9757},
journal = {J. Artif. Int. Res.},
month = mar,
pages = {61–76},
numpages = {16}
}

@inbook{Jordan1999,
author = {Jordan, Michael I. and Ghahramani, Zoubin and Jaakkola, Tommi S. and Saul, Lawrence K.},
title = {An introduction to variational methods for graphical models},
year = {1999},
isbn = {0262600323},
publisher = {MIT Press},
address = {Cambridge, MA, USA},
booktitle = {Learning in Graphical Models},
pages = {105–161},
numpages = {57}
}

@InProceedings{Ranganath14,
  title = 	 {{Black Box Variational Inference}},
  author = 	 {Ranganath, Rajesh and Gerrish, Sean and Blei, David},
  booktitle = 	 {Proceedings of the Seventeenth International Conference on Artificial Intelligence and Statistics},
  pages = 	 {814--822},
  year = 	 {2014},
  editor = 	 {Kaski, Samuel and Corander, Jukka},
  volume = 	 {33},
  series = 	 {Proceedings of Machine Learning Research},
  address = 	 {Reykjavik, Iceland},
  publisher =    {PMLR},
  url = 	 {https://proceedings.mlr.press/v33/ranganath14.html}
}

@article{Hoffman13a,
  author  = {Matthew D. Hoffman and David M. Blei and Chong Wang and John Paisley},
  title   = {Stochastic Variational Inference},
  journal = {Journal of Machine Learning Research},
  year    = {2013},
  volume  = {14},
  number  = {40},
  pages   = {1303--1347},
  url     = {http://jmlr.org/papers/v14/hoffman13a.html}
}

@inproceedings{Kingma13_VAE,
  author       = {Diederik P. Kingma and
                  Max Welling},
  editor       = {Yoshua Bengio and
                  Yann LeCun},
  title        = {Auto-Encoding Variational Bayes},
  booktitle    = {2nd International Conference on Learning Representations, {ICLR} 2014,
                  Banff, AB, Canada, April 14-16, 2014, Conference Track Proceedings},
  year         = {2014},
  url          = {http://arxiv.org/abs/1312.6114},
  bibsource    = {dblp computer science bibliography, https://dblp.org}
}

@article{Gershman2014Amortized,
  title={Amortized Inference in Probabilistic Reasoning},
  author={Samuel J. Gershman and Noah D. Goodman},
  journal={Cognitive Science},
  year={2014},
  volume={36},
  url={https://api.semanticscholar.org/CorpusID:924780}
}

@InProceedings{Omori2013,
author="Omori, Toshiaki",
editor="Lee, Minho
and Hirose, Akira
and Hou, Zeng-Guang
and Kil, Rhee Man",
title="Extracting Latent Dynamics from Multi-dimensional Data by Probabilistic Slow Feature Analysis",
booktitle="Neural Information Processing",
year="2013",
publisher="Springer Berlin Heidelberg",
address="Berlin, Heidelberg",
pages="108--116",
}

@article{Omori2020,
author = {Tsujimoto, Kazuki and Omori, Toshiaki},
year = {2020},
month = {11},
pages = {740-745},
title = {Switching Probabilistic Slow Feature Analysis for Time Series Data},
volume = {10},
journal = {International Journal of Machine Learning and Computing},
doi = {10.18178/ijmlc.2020.10.6.999}
}

@article{Fan2018,
title = {Identification of robust probabilistic slow feature regression model for process data contaminated with outliers},
journal = {Chemometrics and Intelligent Laboratory Systems},
volume = {173},
pages = {1-13},
year = {2018},
issn = {0169-7439},
doi = {https://doi.org/10.1016/j.chemolab.2017.12.009},
url = {https://www.sciencedirect.com/science/article/pii/S0169743917306275},
author = {Lei Fan and Hariprasad Kodamana and Biao Huang}
}

@ARTICLE{Puli2024,
  author={Puli, Vamsi Krishna and Huang, Biao},
  journal={IEEE Transactions on Industrial Informatics}, 
  title={Nonlinear Slow Feature Analysis for Oscillating Characteristics Under Deep Encoder-Decoder Framework}, 
  year={2024},
  volume={20},
  number={7},
  pages={9568-9578},
  doi={10.1109/TII.2024.3383534}}

@article{Roweis1999,
    author = {Roweis, Sam and Ghahramani, Zoubin},
    title = "{A Unifying Review of Linear Gaussian Models}",
    journal = {Neural Computation},
    volume = {11},
    number = {2},
    pages = {305-345},
    year = {1999},
    month = {02},
    issn = {0899-7667},
    doi = {10.1162/089976699300016674},
    url = {https://doi.org/10.1162/089976699300016674},
    eprint = {https://direct.mit.edu/neco/article-pdf/11/2/305/814052/089976699300016674.pdf},
}

@article{Girin2021,
url = {http://dx.doi.org/10.1561/2200000089},
year = {2021},
volume = {15},
journal = {Foundations and Trends® in Machine Learning},
title = {Dynamical Variational Autoencoders: A Comprehensive Review},
doi = {10.1561/2200000089},
issn = {1935-8237},
number = {1-2},
pages = {1-175},
author = {Laurent Girin and Simon Leglaive and Xiaoyu Bie and Julien Diard and Thomas Hueber and Xavier Alameda-Pineda}
}

@article{WiskottSejnowski2002,
author       = {Laurenz Wiskott and Terrence Sejnowski},
title        = {Slow feature analysis: unsupervised learning of invariances.},
journal      = {Neural Computation},
year         = {2002},
volume       = {14},
number       = {4},
pages        = {715--770},
}

@inproceedings{Schueler2019,
  author    = {Merlin Sch{\"{u}}ler and
               Hlynur Dav{\'{\i}}{\dh} Hlynsson and
               Laurenz Wiskott},
  editor    = {Wee Sun Lee and
               Taiji Suzuki},
  title     = {Gradient-based Training of Slow Feature Analysis by Differentiable
               Approximate Whitening},
  booktitle = {Proceedings of The 11th Asian Conference on Machine Learning, {ACML}
               2019, 17-19 November 2019, Nagoya, Japan},
  series    = {Proceedings of Machine Learning Research},
  volume    = {101},
  pages     = {316--331},
  publisher = {{PMLR}},
  year      = {2019},
  url       = {http://proceedings.mlr.press/v101/schuler19a.html},
  bibsource = {dblp computer science bibliography, https://dblp.org}
}

@article{Turner2007,
author = {Turner, Richard and Sahani, Maneesh},
title = {A Maximum-Likelihood Interpretation for Slow Feature Analysis},
year = {2007},
issue_date = {April 2007},
publisher = {MIT Press},
address = {Cambridge, MA, USA},
volume = {19},
number = {4},
issn = {0899-7667},
url = {https://doi.org/10.1162/neco.2007.19.4.1022},
doi = {10.1162/neco.2007.19.4.1022},
journal = {Neural Comput.},
month = apr,
pages = {1022–1038},
numpages = {17}
}

@InProceedings{Wiskott1998,
author="Wiskott, Laurenz",
editor="Niklasson, Lars
and Bod{\'e}n, Mikael
and Ziemke, Tom",
title="Learning Invariance Manifolds",
booktitle="ICANN 98",
year="1998",
publisher="Springer London",
address="London",
pages="555--560",
isbn="978-1-4471-1599-1"
}

@incollection{Bengio2009,
title = {Slow, Decorrelated Features for Pretraining Complex Cell-like Networks},
author = {Bengio, Yoshua and James S. Bergstra},
booktitle = {Advances in Neural Information Processing Systems 22},
editor = {Y. Bengio and D. Schuurmans and J. D. Lafferty and C. K. I. Williams and A. Culotta},
pages = {99--107},
year = {2009},
publisher = {Curran Associates, Inc.},
}

@article{Escalante2013,
 author = {Escalante-B., Alberto N. and Wiskott, Laurenz},
 title = {How to Solve Classification and Regression Problems on High-dimensional Data with a Supervised Extension of Slow Feature Analysis},
 journal = {J. Mach. Learn. Res.},
 issue_date = {January 2013},
 volume = {14},
 number = {1},
 month = dec,
 year = {2013},
 issn = {1532-4435},
 pages = {3683--3719},
 numpages = {37},
 acmid = {2627675},
 publisher = {JMLR.org},
}

@article{Kingma2014,
  author    = {Diederik P. Kingma and
               Jimmy Ba},
  title     = {Adam: {A} Method for Stochastic Optimization},
  journal   = {CoRR},
  volume    = {abs/1412.6980},
  year      = {2014},
  archivePrefix = {arXiv},
  eprint    = {1412.6980},
}

\begin{appendices}
\section{Necessary Conditions for Optimality of Linear Variational SFA}
\label{appendix:necessary_conditions_for_linear_sfavae}
This shows in some detail the derivation from Section \ref{sec:vaesfa_elbo-derivation} of the objective and necessary conditions if one assumes the encoder and decoder, both, to be linear and Gaussian:
\begin{align}
    q(\mathbf{z}|\mathbf{x}) &= \mathcal{N}(\mathbf{z} |\mathbf{W}^T\mathbf{x} + \mathbf{b}, \mathbf{I})\\
    p(\mathbf{x}|\mathbf{z}) &= \mathcal{N}(\mathbf{x} |\mathbf{V}\mathbf{z} + \mathbf{o}, \mathbf{I}).
\end{align}
The derivation is not complicated but length and not particularly insightful, which is why it has been added as appendix.
Plugging the assumptions into the ELBO objective leads to 
\begin{align}
    &\mathcal{L}(\mathbf{x}_{1:T}, \mathbf{W}^T, \mathbf{V}, \mathbf{o}, \mathbf{b})\nonumber\\&=\underbrace{\sum_t\log\left(\frac{1}{\sqrt{2\pi}}\right)}_{\text{const.}\,c} \underbrace{- \frac{1}{2}\sum_t \mathbb{E}_{\varepsilon}\Big[ \|\mathbf{x}_{t} - \big(\mathbf{V}\big(\mathbf{W}^T\mathbf{x}_{t}+\mathbf{b}+ \varepsilon\big) + \mathbf{o} \Big)\|^2\Big]}_\text{A} \underbrace{- \frac{1}{2}\sum_t \|\mathbf{W}^T\mathbf{x}_{t-1} - \mathbf{W}^T\mathbf{x}_t\|^2}_\text{B}
\end{align}
where $\varepsilon \sim \mathcal{N}(\mathbf{0}, \mathbf{I})$. In the following we will simplify and take the derivative w.r.t. the parameters for each part individually.

\paragraph{Part A} 
\begin{align}
&-\frac{1}{2}\sum_t \mathbb{E}_{\varepsilon}\Big[ \|\mathbf{x}_{t} - \big(\mathbf{V}\big(\mathbf{W}^T\mathbf{x}_{t}+\mathbf{b}+ \varepsilon\big) + \mathbf{o} \Big)\|^2\Big]
\\&= -\frac{1}{2}\sum_t \mathbb{E}_{\varepsilon}\Big[ \|\mathbf{x}_{t} - \Big(\mathbf{V}\big(\mathbf{W}^T\mathbf{x}_{t}+\mathbf{b}\big) + \mathbf{o}\Big) - \mathbf{V}\varepsilon \|^2\Big]\\
&= -\frac{1}{2}\sum_t \mathbb{E}_{\varepsilon}\Big[ \| \underbrace{\Big(\mathbf{x}_t -  \Big(\mathbf{V}\big(\mathbf{W}^T\mathbf{x}_{t}+\mathbf{b}\big) + \mathbf{o}\Big)\Big)}_{\mathbf{s}_t} - \mathbf{V}\varepsilon \|^2\Big]
\\
&=-\frac{1}{2} \sum_t \mathbb{E}_{\varepsilon}\Big[ \|\mathbf{s}_t-\mathbf{V}\varepsilon \|^2\Big]
\\
&= -\frac{1}{2}\sum_t \mathbb{E}_{\varepsilon}\Big[ \mathbf{s}_t^T\mathbf{s}_t - 2\mathbf{s}_t^T\mathbf{V}\varepsilon + \varepsilon\mathbf{V}^T\mathbf{V}\varepsilon \Big]
\\
&= -\frac{1}{2}\sum_t \Big[ \mathbb{E}_{\varepsilon}\Big[ \mathbf{s}_t^T\mathbf{s}_t\Big] - 2\mathbb{E}_{\varepsilon}\Big[\mathbf{s}_t^T\mathbf{V}\varepsilon\Big] + \mathbb{E}_{\varepsilon}\Big[\varepsilon\mathbf{V}^T\mathbf{V}\varepsilon \Big]\Big]
\\
&=-\frac{1}{2}\sum_t \Big[ \underbrace{\mathbb{E}_{\varepsilon}\Big[ \mathbf{s}_t^T\mathbf{s}_t\Big]}_{\mathbf{s}_t^T\mathbf{s}_t}- 2\mathbf{s}_t^T\mathbf{V}\underbrace{\mathbb{E}_{\varepsilon}\Big[\varepsilon\Big]}_{\mathbf{0}} + \Tr (\mathbf{V}^T\mathbf{V}\underbrace{\mathbb{E}_{\varepsilon}\Big[\varepsilon\varepsilon^T\Big]}_{\mathbf{I}})\Big]
\\
&=-\frac{1}{2}\sum_t\Big[\mathbf{s}_t^T\mathbf{s}_t + \Tr (\mathbf{V}^T\mathbf{V})\Big]
\\
&=-\frac{1}{2}\sum_t\Big[\|\mathbf{x}_{t} - \Big(\mathbf{V}\big(\mathbf{W}^T\mathbf{x}_{t}+\mathbf{b}\big) + \mathbf{o}\Big)\|^2 + \Tr (\mathbf{V}^T\mathbf{V})\Big]
\end{align}
This can be understood as MSE loss with L2 regularization on the entries of $\mathbf{V}$, since $\Tr (\mathbf{V}^T\mathbf{V}) = \sum_{ij} V_{ij}^2$.
If the norm is now unpacked in to single terms:
\begin{align}
    &-\frac{1}{2}\sum_t\Big[\|\mathbf{x}_{t} - \Big(\mathbf{V}\big(\mathbf{W}^T\mathbf{x}_{t}+\mathbf{b}\big) + \mathbf{o}\Big)\|^2 + \Tr (\mathbf{V}^T\mathbf{V})\Big]
    \\&=-\frac{1}{2} \sum_t \Big[ \mathbf{x}_t^T\mathbf{x}_t - 2\mathbf{x}_t^T\mathbf{VW}^T\mathbf{x}_t - 2\mathbf{x}_t^T\mathbf{Vb} - 2\mathbf{x}_t^T\mathbf{o} + \mathbf{x}_t^T\mathbf{WV}^T\mathbf{V}W\mathbf{x}_t \\&+ 2\mathbf{x}_t^T\mathbf{WV}^T\mathbf{Vb} +2 \mathbf{x}_t^T\mathbf{WV}^T\mathbf{o} + \mathbf{b}^T\mathbf{V}^T\mathbf{Vb} + 2 \mathbf{b}^T\mathbf{V}^T\mathbf{o} + \mathbf{o}^T\mathbf{o} + \Tr(\mathbf{V}^T\mathbf{V})\Big]
    \intertext{Applying the trace and rearranging}
    \\&=-\frac{1}{2} \sum_t \Big[\Tr(\mathbf{x}_t\mathbf{x}_t^T) - 2\Tr(\mathbf{VW}^T\mathbf{x}_t\mathbf{x}_t^T) - 2\Tr(\mathbf{Vb}\mathbf{x}_t^T) - 2\mathbf{x}_t^T\mathbf{o} + \Tr(\mathbf{WV}^T\mathbf{V}W\mathbf{x}_t\mathbf{x}_t^T) \\&+ 2\Tr(\mathbf{WV}^T\mathbf{Vb}\mathbf{x}_t^T) +2 \Tr(\mathbf{WV}^T\mathbf{o}\mathbf{x}_t^T) + \Tr(\mathbf{V}^T\mathbf{Vb}\mathbf{b}^T) + 2\Tr(\mathbf{V}^T\mathbf{o}\mathbf{b}^T) + \mathbf{o}^T\mathbf{o} + \Tr(\mathbf{V}^T\mathbf{V})\Big]
    \intertext{Trace and sum are linear operators and can be interchanged. Using $\mathbf{C}=\frac{1}{T}\sum_t \mathbf{xx}^T$ and $\bar{\mathbf{x}})=\frac{1}{T}\sum_t \mathbf{x}_t$:}
    \\&=-\frac{T}{2}\Big[\Tr(\mathbf{C}) - 2\Tr(\mathbf{VW}^T\mathbf{C}) - 2\Tr(\mathbf{Vb}\bar{\mathbf{x}}^T) - 2\bar{\mathbf{x}}^T\mathbf{o} + \Tr(\mathbf{WV}^T\mathbf{VW}^T\mathbf{C}) \\&+ 2\Tr(\mathbf{WV}^T\mathbf{Vb}\bar{\mathbf{x}}^T) +2\Tr(\mathbf{WV}^T\mathbf{o}\bar{\mathbf{x}}^T) + \Tr(\mathbf{V}^T\mathbf{Vb}\mathbf{b}^T) + 2\Tr(\mathbf{V}^T\mathbf{o}\mathbf{b}^T) + \mathbf{o}^T\mathbf{o} + \Tr(\mathbf{V}^T\mathbf{V})\Big]
\end{align}
We can use these single terms to simplify the derivatives later on.

\paragraph{Part B}
\begin{align}
    &-\frac{1}{2}\sum_t \|\mathbf{W}^T\mathbf{x}_{t-1}-\mathbf{W}^T\mathbf{x}_t\|^2\\
    &=-\frac{1}{2}\sum_t \Big[\mathbf{x}_{t-1}^T\mathbf{WW^T}\mathbf{x}_{t-1} - 2 \mathbf{x}_{t}^T\mathbf{WW^T}\mathbf{x}_{t-1}+\mathbf{x}_{t}^T\mathbf{WW^T}\mathbf{x}_{t}\Big]\\
    &=-\frac{1}{2}\sum_t \Big[\Tr(\mathbf{WW^T}\mathbf{x}_{t-1}\mathbf{x}_{t-1}^T) - 2\Tr(\mathbf{WW^T}\mathbf{x}_{t-1}\mathbf{x}_{t}^T)+\Tr(\mathbf{WW^T}\mathbf{x}_{t}\mathbf{x}_{t}^T)\Big]\\
    \intertext{For large $T$, assume $\mathbf{C}=\frac{1}{T}\sum_t \mathbf{x}_{t-1}\mathbf{x}_{t-1}^T$. Also, we denote $\mathbf{C}_{t;t-1}=\frac{1}{T}\sum_t\mathbf{x}_t\mathbf{x}_{t-1}^T$.}
    &=-\frac{T}{2}\Big[\Tr(\mathbf{WW^T}\mathbf{C}) - 2\Tr(\mathbf{WW^T}\mathbf{C}_{t;t-1})+\Tr(\mathbf{WW^T}\mathbf{C})\Big]\\
    &=-\frac{T}{2}\Big[2\Tr(\mathbf{WW^T}\mathbf{C}) - 2\Tr(\mathbf{WW^T}\mathbf{C}_{t;t-1})\Big]\\
\end{align}

\paragraph{Derivatives of A and B}
The following table lists the derivatives of each term in the sum for more convenient display. Multiplying each with each with $-\frac{T}{2}$ and summing them will give the partial derivative of $\mathcal{L}$ w.r.t. to the corresponding parameters.
\begin{center}
\begin{tabular}{|c|c|c|c|c| } 
\cline{2-5}
 \multicolumn{1}{c|}{} & $\frac{\partial}{\partial \mathbf{V}}$ & $\frac{\partial}{\partial \mathbf{W}^T}$ & $\frac{\partial}{\partial \mathbf{o}}$ & $\frac{\partial}{\partial \mathbf{b}}$   \\ 
  \hline
  \multicolumn{5}{|l|}{\textbf{Part A}} \\
  \hline
 $\Tr(\mathbf{C})$ & $0$ & $0$ & $0$ & $0$\\
 $-2\Tr(\mathbf{VW}^T\mathbf{C})$ & $-2\mathbf{CW}$ & $-2\mathbf{V}^T\mathbf{C}$ & $0$ & $0$\\ 
 $-2\Tr(\mathbf{Vb\bar{x}}^T)$ & $-2\mathbf{\bar{\mathbf{x}}b}^T$ & $0$ & $0$ & $-2\mathbf{\bar{x}}^T\mathbf{V}$\\ 
 $-2\mathbf{\bar{x}}^T\mathbf{o}$ & $0$ & $0$ & $-2\mathbf{\bar{x}}^T$ & $0$\\ 
  $\Tr(\mathbf{WV}^T\mathbf{VW}^T\mathbf{C})$ & $2\mathbf{VW}^T\mathbf{CW}$ & $2\mathbf{V}^T\mathbf{VW}^T\mathbf{C}$ & $0$ & $0$\\ 
  $2\Tr(\mathbf{WV}^T\mathbf{Vb}\bar{\mathbf{x}}^T)$ & $2\mathbf{V}\mathbf{b}\bar{\mathbf{x}}^T\mathbf{W}+\mathbf{VW}^T\bar{\mathbf{x}}\mathbf{b}^T$ & $2\mathbf{V}^T\mathbf{Vb\bar{x}}^T$ & $0$ & $2\bar{\mathbf{x}}^T\mathbf{WV}^T\mathbf{V}$\\ 
  $2\Tr(\mathbf{WV}^T\mathbf{o}\bar{\mathbf{x}}^T)$ & $2\mathbf{o\bar{x}}^T\mathbf{W}$ & $2\mathbf{V}^T\mathbf{o\bar{x}}^T$ & $2\mathbf{\bar{x}}^T\mathbf{WV}^T$ & $0$\\ 
  $\Tr(\mathbf{V}^T\mathbf{Vb}\mathbf{b}^T)$ & $2\mathbf{Vbb}^T$ & $0$ & $0$ & $2\mathbf{b}^T\mathbf{V}^T\mathbf{V}$\\ 
  $2\Tr(\mathbf{V}^T\mathbf{o}\mathbf{b}^T)$ & $2\mathbf{ob}^T$ & $0$ & $2\mathbf{b}^T\mathbf{V}^T$ & $2\mathbf{o}^T\mathbf{V}$\\ 
  $\mathbf{o}^T\mathbf{o} $ & $0$ & $0$ & $2\mathbf{o}^T$ & $0$\\ 
  $\Tr(\mathbf{V}^T\mathbf{V})$ & $2\mathbf{V}$ & $0$ & $0$ & $0$\\ 
 \hline
  \multicolumn{5}{|l|}{\textbf{Part B}} \\
  \hline
  $2\Tr(\mathbf{WW^T}\mathbf{C})$ & $0$ & $4\mathbf{W}^T\mathbf{C}$ & $0$ & $0$\\ 
  $-2\Tr(\mathbf{WW^T}\mathbf{C}_{t;t-1})$ & $0$ & $-4\mathbf{W}^T\mathbf{C}+2\mathbf{W}^T\dot{\mathbf{C}}$ & $0$ & $0$ \\
  \hline
\end{tabular}
\end{center}
Considering first the necessary optimality conditions on $\mathbf{b}$ and $\mathbf{o}$. We take the liberty to overload $\mathbf{0}$ with the zero-vector or matrix with the dimensions corresponding to the context.
\begin{align}
    &\frac{\partial \mathcal{L}}{\partial \mathbf{\mathbf{b}}} = \mathbf{0}\quad\land\quad
    \frac{\partial \mathcal{L}}{\partial \mathbf{\mathbf{o}}} = \mathbf{0} \\
    \Leftrightarrow\quad&
    T\big(\mathbf{\bar{x}}^T - \mathbf{\bar{x}}^T\mathbf{WV}^T - \mathbf{b}^T\mathbf{V}^T - \mathbf{o}^T\big) = \mathbf{0}\\
    \land\quad&T\big( \mathbf{\bar{x}}^T\mathbf{V} - \mathbf{\bar{x}}^T\mathbf{WV}^T\mathbf{V} - \mathbf{b}^T\mathbf{V}^T\mathbf{V} - \mathbf{o}^T\mathbf{V}\big) = \mathbf{0}\\
    \Leftrightarrow\quad&\mathbf{\bar{x}}^T - \mathbf{\bar{x}}^T\mathbf{WV}^T - \mathbf{b}^T\mathbf{V}^T - \mathbf{o}^T = \mathbf{0}\\
    \land\quad&\mathbf{\bar{x}}^T\mathbf{V} - \mathbf{\bar{x}}^T\mathbf{WV}^T\mathbf{V} - \mathbf{b}^T\mathbf{V}^T\mathbf{V} - \mathbf{o}^T\mathbf{V} = \mathbf{0}\\
    \Leftrightarrow\quad& \mathbf{o}^T =\mathbf{\bar{x}}^T - \mathbf{\bar{x}}^T\mathbf{WV}^T - \mathbf{b}^T\mathbf{V}^T\\
    \land\quad&\mathbf{\bar{x}}^T\mathbf{V} - \mathbf{\bar{x}}^T\mathbf{WV}^T\mathbf{V} - \mathbf{b}^T\mathbf{V}^T\mathbf{V} - \mathbf{o}^T\mathbf{V} = \mathbf{0}
    \intertext{Plugging $\mathbf{o}^T$ into the second condition:}
    \Leftrightarrow\quad& \mathbf{o}^T =\mathbf{\bar{x}}^T - \mathbf{\bar{x}}^T\mathbf{WV}^T - \mathbf{b}^T\mathbf{V}^T
    \\\land\quad&\mathbf{\bar{x}}^T\mathbf{V} - \mathbf{\bar{x}}^T\mathbf{WV}^T\mathbf{V} - \mathbf{b}^T\mathbf{V}^T\mathbf{V} - \big(\mathbf{\bar{x}}^T - \mathbf{\bar{x}}^T\mathbf{WV}^T - \mathbf{b}^T\mathbf{V}^T\big)\mathbf{V} = \mathbf{0}\\
    \Leftrightarrow\quad& \mathbf{o}^T =\mathbf{\bar{x}}^T - \mathbf{\bar{x}}^T\mathbf{WV}^T - \mathbf{b}^T\mathbf{V}^T
    \quad\land\quad \mathbf{0} = \mathbf{0}\\
    \Leftrightarrow\quad& \mathbf{o}^T =\mathbf{\bar{x}}^T - \mathbf{\bar{x}}^T\mathbf{WV}^T - \mathbf{b}^T\mathbf{V}^T \label{eq:sfavae_linear_necessary_o}
\end{align}
So $\mathbf{o}$ is determined by the data mean $\mathbf{\bar{x}}$ and the other parameters.
The necessary conditions for $\mathbf{V}$ and $\mathbf{W}^T$ are:
\begin{align}
    &\frac{\partial \mathcal{L}}{\partial \mathbf{V}}=\mathbf{0} \quad\land\quad \frac{\partial \mathcal{L}}{\partial \mathbf{W}^T} = \mathbf{0}
    \\\Leftrightarrow\quad&\nonumber
    T\big(\mathbf{CW} + \mathbf{\bar{x}b}^T - \mathbf{VW}^T\mathbf{CW} - \mathbf{V}\mathbf{b}\bar{\mathbf{x}}^T\mathbf{W}-\mathbf{VW}^T\bar{\mathbf{x}}\mathbf{b}^T 
    \\&-\mathbf{o\bar{x}}^T\mathbf{W} - \mathbf{Vbb}^T - \mathbf{ob}^T - \mathbf{V}\big) = \mathbf{0}
    \\\land\quad&
    T\big(\mathbf{V}^T\mathbf{C} - \mathbf{V}^T\mathbf{VW}^T\mathbf{C} - \mathbf{V}^T\mathbf{V}\mathbf{b\bar{x}}^T-\mathbf{V}^T\mathbf{o\bar{x}}^T \underbrace{- 2\mathbf{W}^T\mathbf{C} + 2\mathbf{W}^T\mathbf{C}}_{\mathbf{0}}-\mathbf{W}^T\mathbf{\dot{C}}\big) = \mathbf{0}
    \\\Leftrightarrow\quad&\nonumber
    \mathbf{CW} + \mathbf{\bar{x}b}^T - \mathbf{VW}^T\mathbf{CW} - \mathbf{V}\mathbf{b}\bar{\mathbf{x}}^T\mathbf{W}-\mathbf{VW}^T\bar{\mathbf{x}}\mathbf{b}^T \\&-\mathbf{o\bar{x}}^T\mathbf{W}- \mathbf{Vbb}^T - \mathbf{ob}^T - \mathbf{V} = \mathbf{0}
    \\\land\quad&
    \mathbf{V}^T\mathbf{C} - \mathbf{V}^T\mathbf{VW}^T\mathbf{C} - \mathbf{V}^T\mathbf{V}\mathbf{b\bar{x}}^T-\mathbf{V}^T\mathbf{o\bar{x}}^T-\mathbf{W}^T\mathbf{\dot{C}} = \mathbf{0}
    \intertext{Multiplying the first line from the left with $\mathbf{V}^T$ and the second line from the right with $\mathbf{W}$ and underlining terms that occur in both:}
    \Leftrightarrow\quad&\nonumber
    \underline{\mathbf{V}^T\mathbf{CW}} + \mathbf{V}^T\mathbf{\bar{x}b}^T - \underline{\mathbf{V}^T\mathbf{VW}^T\mathbf{CW}} - \mathbf{V}^T\mathbf{V}\mathbf{b}\bar{\mathbf{x}}^T\mathbf{W}-\mathbf{V}^T\mathbf{VW}^T\bar{\mathbf{x}}\mathbf{b}^T \\&-\underline{\mathbf{V}^T\mathbf{o\bar{x}}^T\mathbf{W}}- \mathbf{V}^T\mathbf{Vbb}^T - \mathbf{V}^T\mathbf{ob}^T - \mathbf{V}^T\mathbf{V} = \mathbf{0}
    \\\land\quad&
    \underline{\mathbf{V}^T\mathbf{C}\mathbf{W}} - \underline{\mathbf{V}^T\mathbf{VW}^T\mathbf{C}\mathbf{W}} - \mathbf{V}^T\mathbf{V}\mathbf{b\bar{x}}^T\mathbf{W}-\underline{\mathbf{V}^T\mathbf{o\bar{x}}^T\mathbf{W}}-\mathbf{W}^T\mathbf{\dot{C}}\mathbf{W} = \mathbf{0}
    \intertext{Combining both lines on equal terms:}
    \Leftrightarrow\quad& \nonumber
        \mathbf{V}^T\mathbf{\bar{x}b}^T - \mathbf{V}^T\mathbf{V}\mathbf{b}\bar{\mathbf{x}}^T\mathbf{W}-\mathbf{V}^T\mathbf{VW}^T\bar{\mathbf{x}}\mathbf{b}^T  - \mathbf{V}^T\mathbf{Vbb}^T - \mathbf{V}^T\mathbf{ob}^T - \mathbf{V}^T\mathbf{V}
        \\&= - \mathbf{V}^T\mathbf{V}\mathbf{b\bar{x}}^T\mathbf{W}-\mathbf{W}^T\mathbf{\dot{C}}\mathbf{W}
    \\\Leftrightarrow\quad&
        \mathbf{V}^T\mathbf{\bar{x}b}^T -\mathbf{V}^T\mathbf{VW}^T\bar{\mathbf{x}}\mathbf{b}^T  - \mathbf{V}^T\mathbf{Vbb}^T - \mathbf{V}^T\mathbf{ob}^T - \mathbf{V}^T\mathbf{V}
        = -\mathbf{W}^T\mathbf{\dot{C}}\mathbf{W}
    \intertext{Plugging in eq. \ref{eq:sfavae_linear_necessary_o}}
    \Leftrightarrow\quad& \nonumber
        \mathbf{V}^T\mathbf{\bar{x}b}^T -\mathbf{V}^T\mathbf{VW}^T\bar{\mathbf{x}}\mathbf{b}^T  - \mathbf{V}^T\mathbf{Vbb}^T 
        \\&- \mathbf{V}^T\mathbf{\bar{x}b}^T
        +\mathbf{V}^T\mathbf{VW}^T\mathbf{\bar{x}b}^T
        +\mathbf{V}^T\mathbf{V}\mathbf{bb}^T
        - \mathbf{V}^T\mathbf{V}
        = -\mathbf{W}^T\mathbf{\dot{C}}\mathbf{W}
        \\    \Leftrightarrow\quad& 
        \mathbf{V}^T\mathbf{V}
        = \mathbf{W}^T\mathbf{\dot{C}}\mathbf{W}
\end{align}
So the necessary conditions for optimality w.r.t. to the parameters $\mathbf{W}^T, \mathbf{V}, \mathbf{b}, \mathbf{o}$ are 
\begin{align}
        \mathbf{V}^T\mathbf{V} = \mathbf{W}^T\mathbf{\dot{C}}\mathbf{W} \quad \text{and} \quad \mathbf{o}^T =\mathbf{\bar{x}}^T - \mathbf{\bar{x}}^T\mathbf{WV}^T - \mathbf{b}^T\mathbf{V}^T
        \label{eq:sfavae_necessary_conditions_final}
\end{align}
\end{appendices}
\end{document}